\documentclass[sigconf]{acmart}

\acmConference{}{}{}               
\acmBooktitle{}                   
\acmPrice{}                       
\acmISBN{}                         
\acmDOI{}                          

\renewcommand\footnotetextcopyrightpermission[1]{}

\title{Knowledge-based learning in Text-RAG and
Image-RAG}

\author{Alexander Shim}
\affiliation{
  \institution{Florida International University}
  \city{Miami}
  \state{FL}
  \country{USA}
}
\email{ashim017@fiu.edu}

\author{Khalil Saieh}
\affiliation{
  \institution{Florida International University}
  \city{Miami}
  \state{FL}
  \country{USA}
}
\email{ksaie001@fiu.edu}

\author{Samuel Clarke}
\affiliation{
  \institution{Florida International University}
  \city{Miami}
  \state{FL}
  \country{USA}
}
\email{sclar122@fiu.edu}

\newpage

\begin{document}
\maketitle

\section{Abstract}

This research analyzed and compared the multi-modal approach in the Vision Transformer(EVA-ViT) based image encoder with the LlaMA or ChatGPT LLM to reduce the hallucination problem and detect diseases in chest x-ray images. In this research, we utilized the NIH Chest X-ray image to train the model and compared it in image-based RAG, text-based RAG, and baseline. \cite{khan2025chestgpt} \cite{quanyang2024artificial}

In a result, the text-based RAG\cite{aydin2025openai} effectively reduces the hallucination problem by using external knowledge information, and the image-based RAG improved the prediction confidence and calibration by using the KNN methods. \cite{ostrovsky2025evaluating} Moreover, the GPT LLM showed better performance, a low hallucination rate, and better Expected Calibration Error(ECE) than Llama Llama-based model. 

This research shows the challenge of data imbalance, a complex multi-stage structure, but suggests a large experience environment and a balanced example of use. 

\section{Introduction}

Radiology sits at a pivotal moment. While imaging technologies have advanced rapidly, the ability to interpret these images has not scaled at the same pace. Hospitals now produce far more chest X-rays than radiologists can review in a timely manner, and this mismatch contributes to diagnostic delays and rising workload pressures. Becker’s Hospital Review reports that imaging volumes are growing by as much as 5\% annually, while radiology residency slots expand by only about 2\% each year. Rather than viewing this challenge as a simple shortage of clinical personnel, it can be understood as a signal that traditional image-interpretation pipelines need rethinking.

One promising direction is the development of multi-modal AI systems that treat medical images not as isolated visual artifacts but as sources of information that can be reasoned about linguistically. This shift has been made possible by the emergence of two influential model families. Vision transformers have proven highly effective at extracting rich, structured features from radiographs, while large language models have shown that text-based reasoning can be applied to complex clinical problems. When combined, these models create an opportunity to unify visual perception with medical interpretation in a single computational process.

Yet existing attempts to merge these capabilities often fall short of clinical needs. Many systems excel at either identifying the presence of a disease or describing it in text, but struggle to pinpoint where abnormalities lie within the image. Others provide useful predictions but lack transparency, making it difficult for clinicians to trust the underlying reasoning. A framework that simultaneously classifies conditions, locates relevant findings, and communicates its reasoning clearly would represent a meaningful step forward.

In response to this need, we conduct an analysis inspired by recent multi-modal frameworks such as ChestGPT to evaluate how different retrieval-augmented generation (RAG) strategies perform in chest X-ray interpretation. Rather than introducing a new model, our work focuses on comparing text-based RAG, image-based RAG, and a baseline multi-modal system to determine which approach provides stronger diagnostic support. Using the NIH lung x-ray dataset, we assess each method’s ability to classify disease findings, enabling a systematic comparison of performance and interpretability across RAG modalities.

\begin{figure*}
    \centering
    \includegraphics[width=\textwidth]{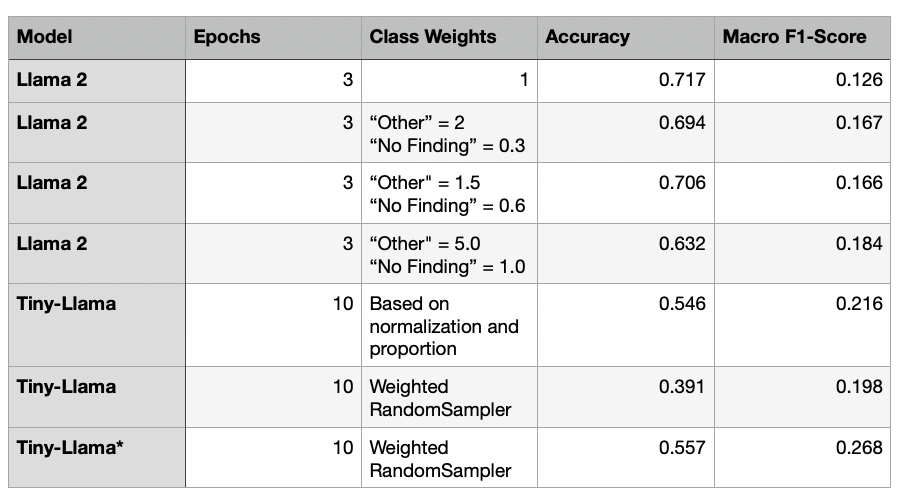}
    \caption{Initial Results (Removed Pneumonia class data from train and test set)}
\end{figure*}

\section{Methodology}

The research is composed of 4 sections: (1) Pre-processing, (2) ChatGPT-based model, (3) LlaMa-based Model, (4)RAG structure of text rag and image rag, (5) learning and strategy of evaluation. In each section, it is explained further in the following sections. 

We used the 112,000 images in the original NIH dataset. After we used single-label filtering, the dataset is 84,053. We separated a total of 84,053 for the training set (80\%), the test set (10\%), and the validation set(10\%). We deleted the pneumonia because it was one of the rarest cases of disease we had to find. The main problem was based on the unbalanced dataset. 

\begin{figure*}
    \centering
    \includegraphics[width=\textwidth]{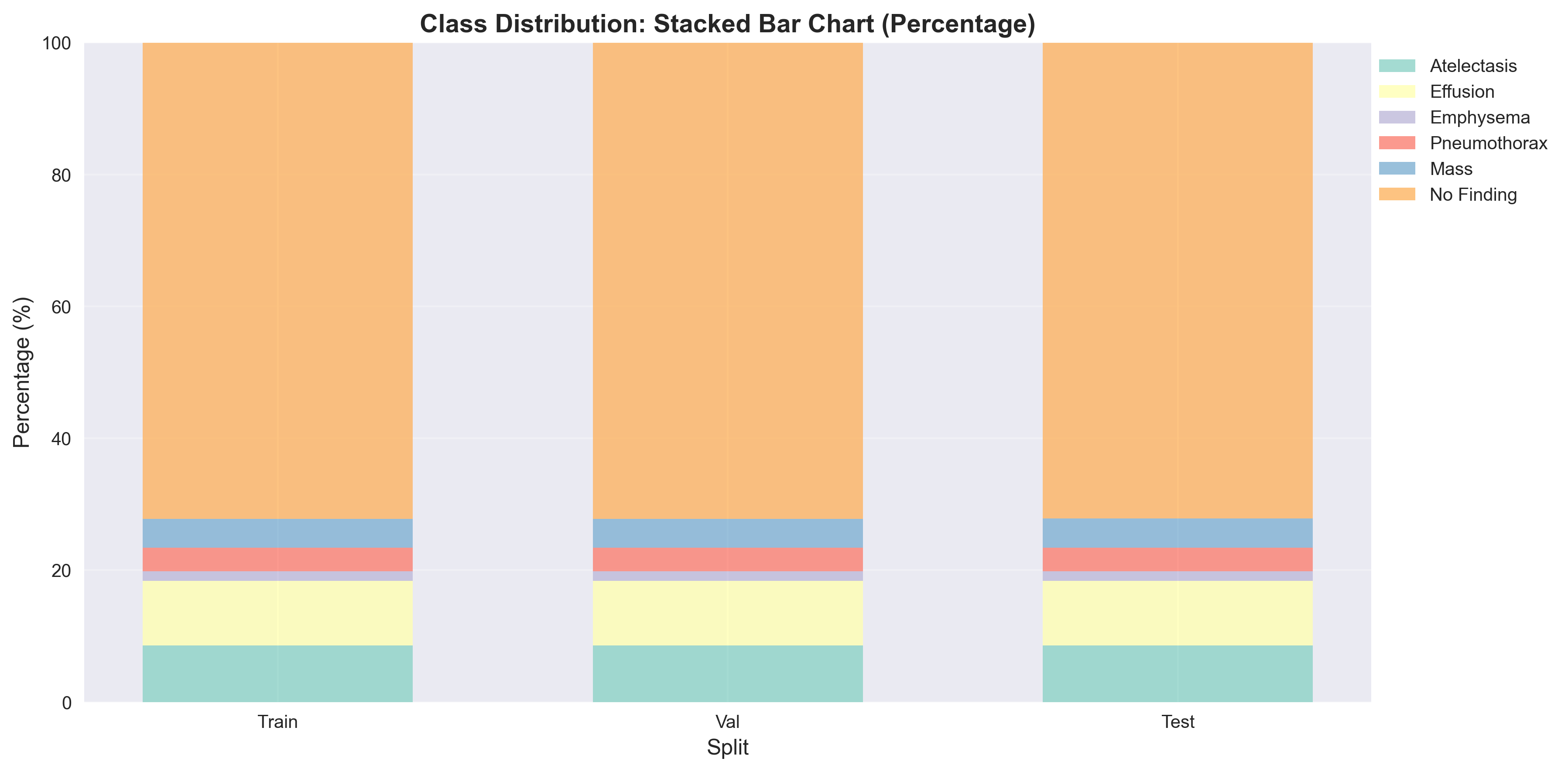}
    \caption{Class Distribution}
\end{figure*}

The problem of class imbalance in Figure 2 is certainly shows that the "No Finding" class dominates over the other classes, and other rare classes were about the diseases. 
\begin{figure*}[t]
    \centering
    \includegraphics[width=0.6\textwidth]{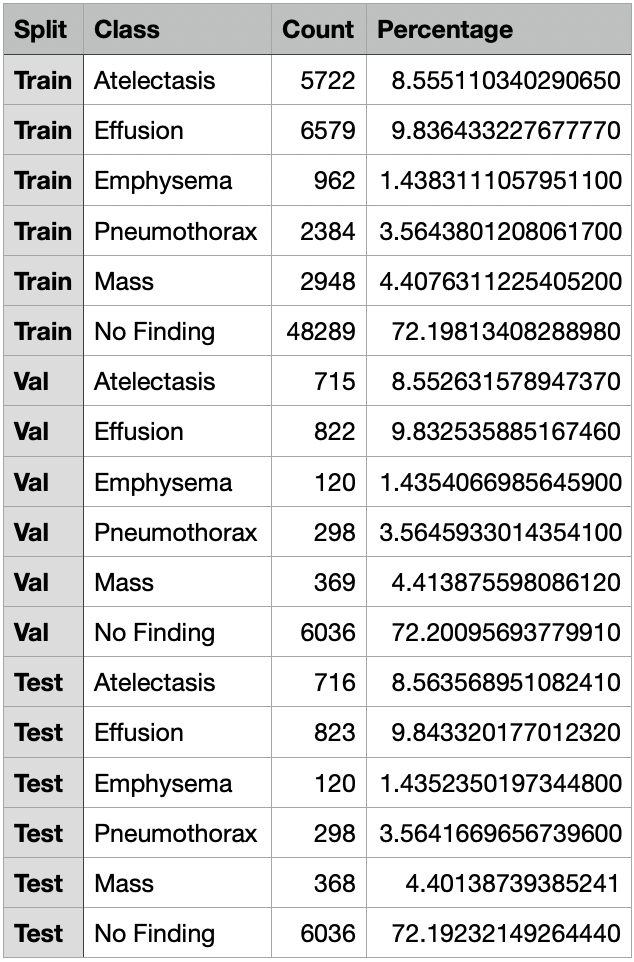}
    \caption{Class Distribution}
\end{figure*}

\begin{figure*}
    \centering
    \includegraphics[width=\textwidth]{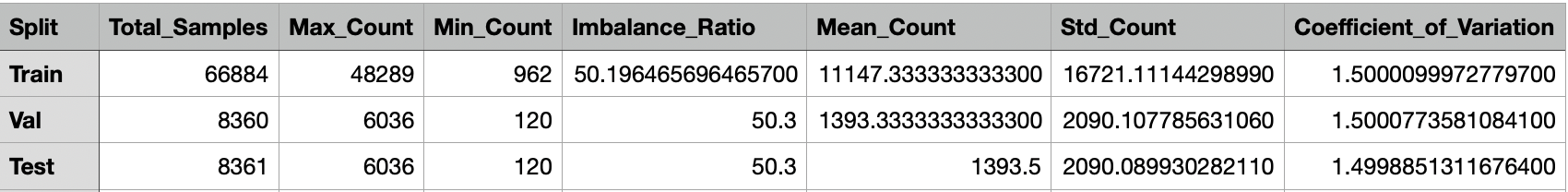}
    \caption{Imbalance ratio}
\end{figure*}

With Figure 3, the class distribution strictly shows that the class of "No Finding" count is much higher than other classes, which causes the imbalanced dataset problem. With the figure 4, the imbalance ratio shows around 50\% in all the training, validation, and testing sets, which is highly skewed. It makes the model exposed to the same skew at every stage of learning and evaluation.

We decided to apply the WeightedRandomSampler than the normal randomsampler to solve several class imbalance problems in the NIH dataset. The normal oversampling method just increases the rate of data continuously and copies it to increase the frequency. However, it shows that the same rare classes overlap excessively and cause the noise, which can lead to the problem of overfitting. However, the undersampling method is deleting some of the dominant classes and making the overall distribution balanced. It might lose the normal sample, which is important, and decrease the variability in the total dataset. 
With two different approaches, we decided to use the WeightedRandomSampler, using each class ratio with sample selection probability, and not delete or copy any data. By using WeightedRandomSampler, we could increase the selection probability for rare classes and decrease the dominant class probability, it gives a balanced gradient update with better training results with rare classes in cross-entropy loss.

\subsection{Pre-processing}

The NIH lung x-ray dataset exhibited a substantial class imbalance, with certain diagnostic categories, particularly “Pneumonia", having extremely few samples relative to more common classes such as “No Finding”. This imbalance negatively impacted model performance, often causing the classifiers to over-predict the majority class and yield near-zero F1 scores for minority classes. To address this, we applied several strategies during preprocessing and training. First, we experimented with class-weighted loss functions, adjusting weights for rare labels to increase their contribution during optimization. Different weighting schemes were tested, including elevating the weights of minority disease labels and down-weighting the dominant “No Finding” class to counteract its disproportionate influence. Additionally, we used a weighted RandomSampler to oversample underrepresented categories, ensuring that each batch contained a more balanced label distribution. Because the “Pneumonia” category had extremely limited representation, we ultimately removed it from the training set to prevent instability and misleading gradients. These interventions collectively aimed to mitigate the imbalance problem; however, as observed in our results, increasing class weights consistently improved macro-F1 scores while reducing accuracy, highlighting the trade-off between balanced performance across classes and overall prediction correctness.
\subsection{ChatGPT-based Model}

In this project, one of the LLMs that we used was GPT. Because of the limited resources that we had, financially and hardware-wise, we were unable to run the full GPT model. What we used instead was GPT mini. It is a smaller model of GPT that made running this project feasible. 

The GPT model was trained from a dataset that had 112,000 records, which came from the NIH. The amount of datasets that were actually used were used was 84,053 records.

We had 3 conditions for each model that we ran (see \textbf{Figure 9}). The models are the baseline, image-based rag, and text-based rag. An image example of this can be found in Figure 5. Similar to the LlaMa model that we will talk about later, we ran the model 20 times for each of the three conditions.

We have provided example results for the following categories: accuracy over epochs (see \textbf{Figure 9}) and hallucination over epochs (see \textbf{Figure 10}). The accuracy over epochs shows that the baseline and text Rag models were the most consistently accurate ones. Even though the text Rag stayed the most consistent, both of them essentially have the same properties. They both stayed in the same 0.15 difference from the lowest to the highest points on the chart. For the hallucination over epochs, the image Rag seems to hallucinate fewer times per epoch compared to the baseline and text Rag. One major important thing to observe is that when the baseline and text Rag hallucinate, they seem to hallucinate a lot. 

\subsection{Llama-based Model}

\begin{figure*}
    \centering
    \includegraphics[width=\textwidth]{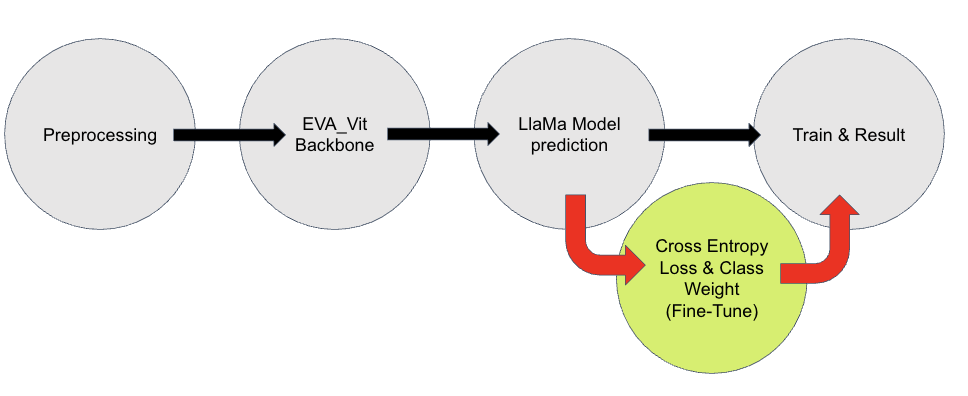}
    \caption{Model design of the LlaMA-based pipeline (Pre-processing → EVA-ViT → LlaMA Prediction → Loss/Weight → Training).}
\end{figure*}

With the NIH Chest-Xray dataset, the LlaMa-based multi-model composition is composed of 4 steps in the full pipeline: (1) pre-processing, (2) EVA-ViT backbone, (3) Llama prediction layer, (4) loss function and class weight, (5) optimization of Adam Optimizer. It is visually explained in Figure 2 and explained further in this subsection.

The first step of the pipeline is pre-processing. In this step, the model filters the one certain categorical that contains only one data set from the NIH dataset. In this research, we only focused on the single-label classification, so we filtered the multi-label diseases.

The second step of the pipeline is the EVA-ViT backbone. It is the vision transformer-based encoder that tokenizes the input X-ray image to image tokens. It allows the LlaMa model to evaluate the language-based clarification with the 8 input token vectors. 

The third step of pipeline is LlaMa prediction layer. The model uses a fine-tuning method to predict one of 6 disease classes: Atelectasis, Effusion, Emphysema, Pneumothorax, Mass, No Finding in LlaMa2 model. The training of the model is based on the cross-entropy loss and uses the class weight to reduce the large class imbalance problem that is discussed in pre-processing.

The next step of the pipeline shows the cross-entropy to stabilize the accuracy of the model. We chose to use the Adam optimizer over others as the Adam optimizer is a popular adaptive learning rate algorithm with stochastic gradient descent in machine learning. We used weighted cross-entropy loss. It helps to reduce the main class gradient and increase the rare class gradient to introduce the learning of the rare class. 

In the last step, we suggest using the Adam optimizer as it regularizes the unstable gradient because the multi-modal pipeline we used generates different gradient scales in each step. In the multi-modal pipeline, the gradient size and variation are different as the structure of the pipeline is composed of EVA-ViT, Llama Model, and classification. It is efficient to reduce the unstable gradient in multi-modal. Moreover, it is efficient to handle the class imbalance as the Adam optimizer can generate the learning rate automatically and balance the difference between classes.

\section{RAG structure of text-rag and image rag}

\begin{figure*}
    \centering
    \includegraphics[width=\textwidth]{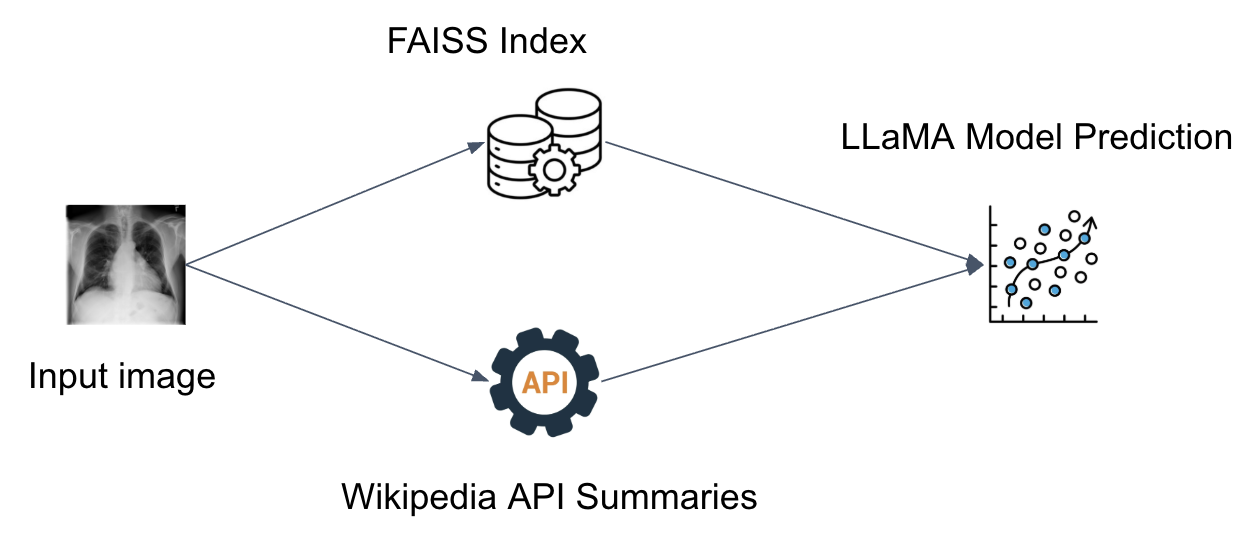}
    \caption{Image RAG vs Text RAG}
\end{figure*}

In this research, we used the RAG to increase the reasoning and prevent the hallucination problem to increase the result of medical video based multi-modal model. RAG is not simply dependent on the input image but predict the reason-grounded with the external information, such as text or image.

For the image rag, the input image is an X-ray image that is a one-labeled disease image. It finds out similar x-ray images with the KNN technique, and our team chose the k value of 3 to efficiently increase the speed of the model. This increases the visual reasoning by creating the FAISS index and search with KNN. As a result, it reduces the false positives and improves the visual consistency. The image-Rag only uses the intra-domain, which is in the data storage. It surely contains the probability of hallucination.

For the text rag, it uses the external medical corpus by referencing the Wikipedia API summaries. Specifically, based on the feature keyword, the related information is searched on an external document and includes the extra natural language context in the model input. It reduces the hallucination and improves the reasoning quality, as the external medical corpus will assist the reasoning behind the prediction of the model. Moreover, it shows the extra domain knowledge, and it is shown as a high level of domain.

\section{learning and strategy of evaluation}

\begin{figure*}
    \centering
    \includegraphics[width=\textwidth]{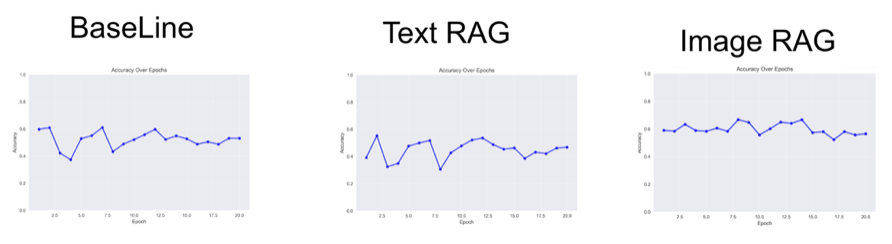}
    \caption{Accuracy over epochs evaluation in Llama Model}
\end{figure*}

\begin{figure*}
    \centering
    \includegraphics[width=\textwidth]{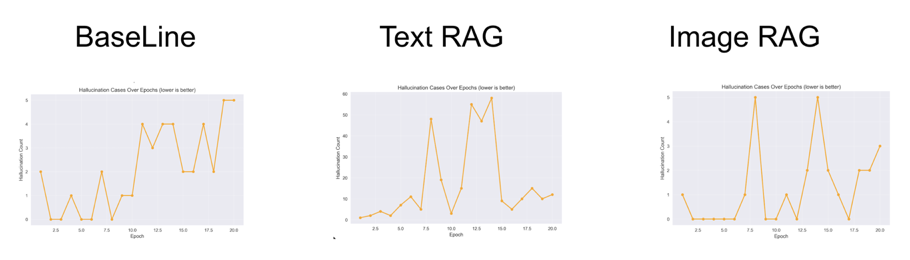}
    \caption{Hallucination over epochs in Llama Model}
\end{figure*}

\begin{figure*}
    \centering
    \includegraphics[width=\textwidth]{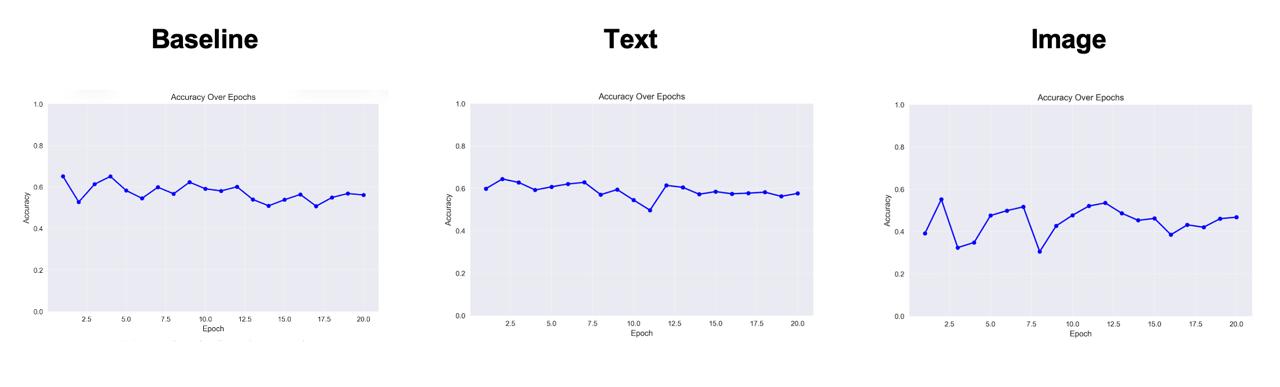}
    \caption{Accuracy over epochs evaluation in ChatGPT Model}
\end{figure*}

\begin{figure*}
    \centering
    \includegraphics[width=\textwidth]{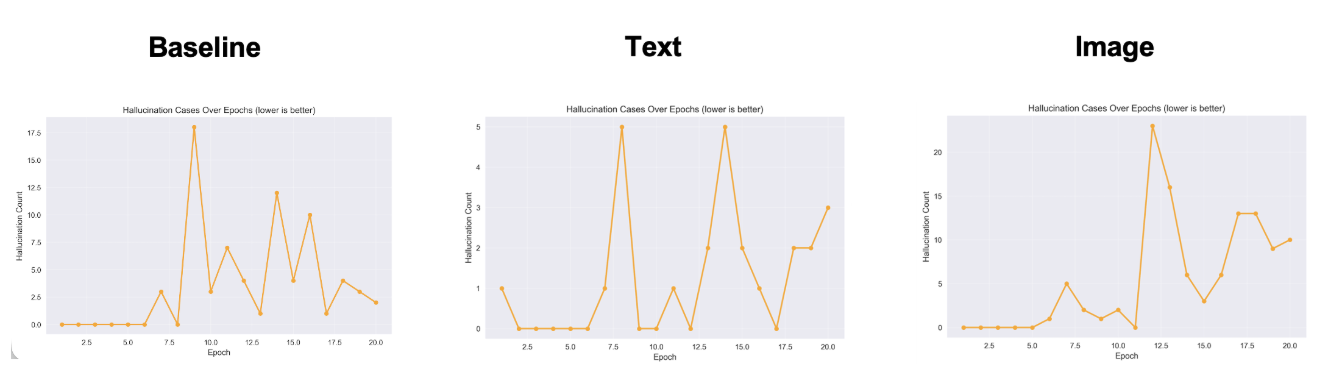}
    \caption{Hallucination over epochs in ChatGPT Model}
\end{figure*}

Baseline model:
The baseline model begins relatively strong (~0.60), then drops sharply (~0.40). The model fluctuates significantly between 0.40–0.60 across the remaining epochs. No clear improvement trend; the model oscillates instead of converging. We interpret that the accuracy curve suggests difficulty learning consistent features from the data without external grounding. The baseline model is likely overfitting or has noisy training dynamics due to limited contextual support.
Text-based RAG model:
In the text-based RAG model, the initial variation is similar to the baseline (around 0.35–0.55). The model shows a steady upward improvement up to ~epoch 12 (reaching ~0.55). There is a slight decline afterward, but it remains more stable than the baseline. We interpret that the RAG with text adds valuable contextual knowledge, helping the model learn more stable decision boundaries. The model peaks slightly below the baseline’s best points, but overall stability is higher, and the mid-epoch improvement is clearer. The slight late-epoch decline may reflect diminishing returns of text retrieval, potential overfitting, or less helpful retrieved passages in later batches.
Image-based RAG Model:
The image-based RAG model has the highest consistency and overall performance. Accuracy stays in the 0.55–0.70 range across nearly all epochs. The trend shows a smooth upward progression early (up to epochs 8–12). There is only mild fluctuations afterward, reaching the highest peak among the three models at ~0.70. We attribute this strong stability to suggest that image-based retrieval anchors the model’s predictions, reducing noise. It shows to provide the best generalization - high accuracy with minimal volatility. The performance of the model forms a clear plateau, ultimately suggesting a sign of stable convergence.

Baseline model:
Early epochs in the baseline model show a hallucination count near 0-2, indicating stable behavior. Later epochs show a gradual increase in hallucination counts around 4-5 by epoch 20. There are no sudden spikes of hallucinations, suggesting the model may be overfitting or losing generalization as training continues. The model lacks grounding information, so once it begins drifting, it keeps drifting higher.
Text-based RAG model:
In the text-based RAG model, earlier epochs show a low hallucination count between 0-10. By epochs 8-14, clear and significant spikes in hallucination are observed between 40-60 hallucinations. After the large spike, the hallucination rate drops again but remains higher than the baseline rate. The trend of hallucinations in the text-based RAG model imply that the model is highly sensitive to retrieved text, indicating that the retrieved text is irrelevant or misleading. During certain epochs, the retrieved text could likely have contradicted the image observed. 
Image-based RAG Model:
In the image-based RAG model, most epochs have a hallucination count within the 0-2 range. There are two distinguished spikes observed, around epochs 8 and 15 where the hallucination counts are near 5. After the spikes, the hallucination rate returns rapidly to near-zero levels. We interpret the trend to imply that the image-based RAG model is the most stable model overall. Visual grounding helps constrain the model’s output. The occasional spikes may correspond to bad batches, retrieval failures, or temporarily unstable gradient updates. However, unlike the text-based RAG model’s performance, the spikes are small and brief.

\section{Results / Discussion}

1) The reason for class weight techniques in pre-processing and LlaMa2 of the prediction layer twice in this research
-> What causes it cause and why we choose it to do -> Results and discussion, what we can do in the next research 

2) The results of comparison in two LLM and two RAGs -> Shows the lack of resources to use the recent model with a large dataset(NIH Chest-xray) 

2.1) Accuracy Over Epochs in Figure 7(Llama Model) and Figure 9(ChatGPT Model)

2.2) Hallucination Cases Over Epochs in Figure 8(Llama Model) and Figure 10(ChatGPT Model)

Baseline model:
The baseline model begins relatively strong (~0.60), then drops sharply (~0.40). The model fluctuates significantly between 0.40–0.60 across the remaining epochs. No clear improvement trend; the model oscillates instead of converging. We interpret that the accuracy curve suggests difficulty learning consistent features from the data without external grounding. The baseline model is likely overfitting or has noisy training dynamics due to limited contextual support.

Text-based RAG model:
In the text-based RAG model, the initial variation is similar to the baseline (around 0.35–0.55). The model shows a steady upward improvement up to ~epoch 12 (reaching ~0.55). There is a slight decline afterward, but it remains more stable than the baseline. We interpret that the RAG with text adds valuable contextual knowledge, helping the model learn more stable decision boundaries. The model peaks slightly below the baseline’s best points, but overall stability is higher, and the mid-epoch improvement is clearer. The slight late-epoch decline may reflect diminishing returns of text retrieval, potential overfitting, or less helpful retrieved passages in later batches.

Image-based RAG Model:
The image-based RAG model has the highest consistency and overall performance. Accuracy stays in the 0.55–0.70 range across nearly all epochs. The trend shows a smooth upward progression early (up to epochs 8–12). There is only mild fluctuations afterward, reaching the highest peak among the three models at ~0.70. We attribute this strong stability to suggest that image-based retrieval anchors the model’s predictions, reducing noise. It shows to provide the best generalization - high accuracy with minimal volatility. The performance of the model forms a clear plateau, ultimately suggesting a sign of stable convergence.

Baseline model:
Early epochs in the baseline model show a hallucination count near 0-2, indicating stable behavior. Later epochs show a gradual increase in hallucination counts around 4-5 by epoch 20. There are no sudden spikes of hallucinations, suggesting the model may be overfitting or losing generalization as training continues. The model lacks grounding information, so once it begins drifting, it keeps drifting higher.

Text-based RAG model:
In the text-based RAG model, earlier epochs show a low hallucination count between 0-10. By epochs 8-14, clear and significant spikes in hallucination are observed between 40-60 hallucinations. After the large spike, the hallucination rate drops again but remains higher than the baseline rate. The trend of hallucinations in the text-based RAG model imply that the model is highly sensitive to retrieved text, indicating that the retrieved text is irrelevant or misleading. During certain epochs, the retrieved text could likely have contradicted the image observed. 

Image-based RAG Model:
In the image-based RAG model, most epochs have a hallucination count within the 0-2 range. There are two distinguished spikes observed, around epochs 8 and 15, where the hallucination counts are near 5. After the spikes, the hallucination rate returns rapidly to near-zero levels. We interpret the trend to imply that the image-based RAG model is the most stable model overall. Visual grounding helps constrain the model’s output. The occasional spikes may correspond to bad batches, retrieval failures, or temporarily unstable gradient updates. However, unlike the text-based RAG model’s performance, the spikes are small and brief.

\section{Conclusions}

Throughout this study, we looked into how LLMs could be integrated within the radiology department. If successful, this would be a huge help in reducing turnaround times for scans done in the department. Faster turnaround times can lead to quicker patient care, at least in this leg of the care. While we want faster turnaround times, we don’t want to compromise the integrity of the results. For this reason, we did initially did two things. The first thing is looking into more efficient ways to improve on the ChestGPT study that was done previously. The second thing we did was document the accuracy over epochs. The model was built, trained, and meant to learn from the previous epochs. This in itself could not guarantee high success rates possible, so we did a few more things. 

One of the main things we did was test two different LLMs to see how they went against each other. The LLMs used were GPT-2 mini and LlaMa 3. These two were tests with a few different key performance indicators. The five that we settled on were the following: accuracy over epochs, hallucination cases over epochs, hallucination by class, expected calibration error over epochs, and calibration plot.

We then compared each key performance indicator against three alterations of each LLM. We first had the baseline so we could have an idea if the alteration improved or weakened the LLM. The two alterations that we added were the text rag and image rag.

The result of all three trials showed the following information. The image rag had the strongest and most stable results, the text rag was not reliable consistently, and the GPT 2 mini performed better than LlaMa 3. Our limitations came mainly from an imbalance NIH dataset and a lack of sufficient hardware. A more balanced dataset and majorly improved hardware would allow us to use a larger version of GPT and run more epochs, potentially improving the results. 

One thing stands true: RAG models can either make the project, like the text RAG, or complicate it, like the image RAG. Future work can definitely improve the results of this trial significantly. 

\bibliographystyle{ACM-Reference-Format}
\bibliography{references}

\end{document}